\documentclass[letterpaper, 10 pt, conference]{ieeeconf}  

\IEEEoverridecommandlockouts                              

\usepackage{stfloats} 

\usepackage{graphics} 
\usepackage{epsfig} 
\usepackage{mathptmx} 
\usepackage{times} 
\usepackage{amsmath} 
\usepackage{amssymb}  

\usepackage{hyperref}
\usepackage{algorithm}
\usepackage{algpseudocode}
\usepackage{multirow}
\usepackage[table]{xcolor}
\usepackage{lipsum}
\usepackage{comment} 
\usepackage{pifont}
\usepackage{listings}
\usepackage{bm}
\usepackage{footmisc}
\usepackage{balance}

\usepackage[backend=biber, style=ieee, maxbibnames=20, minbibnames=20]{biblatex}
\title{\LARGE \bf
A Robotic Skill Learning System Built Upon Diffusion Policies and Foundation Models
}

\author{Nils Ingelhag*, Jesper Munkeby*, Jonne van Haastregt*, Anastasia Varava, Michael C. Welle, Danica Kragic
\thanks{*These authors contributed equally (listed in alphabetical order).}
\thanks{KTH Royal Institute of Technology Stockholm, Sweden, {\it\small \{ingelhag, munkeby, jmvh, mwelle, varava, dani\}@kth.se}}%
}

\begin{document}

\maketitle

\thispagestyle{empty}
\pagestyle{empty}

\begin{abstract}
In this paper, 
we build upon two major recent developments in the field, Diffusion Policies for visuomotor manipulation and large pre-trained multimodal foundational models to obtain a robotic skill learning system.
The system can obtain new skills via the behavioral cloning approach of visuomotor diffusion policies given teleoperated demonstrations. Foundational models are being used to perform skill selection given the user's prompt in natural language.
Before executing a skill the foundational model performs a precondition check given an observation of the workspace.
We compare the performance of different foundational models to this end as well as give a detailed experimental evaluation of the skills taught by the user in simulation and the real world.
Finally, we showcase the combined system on a challenging food serving scenario in the real world.
Videos of all experimental executions, as well as the process of teaching new skills in simulation and the real world, are available on the project's website\footnote{\url{https://roboskillframework.github.io/}\label{fn:website}}.

\end{abstract}

\section{Introduction}
How can we ensure that robots have the necessary skills to accomplish the various tasks that a specific user might need them to do in its specific environment?
While certain skills are potentially more universal than others, the long-tailed nature~\cite{zhang2023deep} of the necessary skills is a major challenge to overcome to make autonomous agents truly ubiquitous in everyday environments.

\begin{figure}[t]
    \centering
    \includegraphics[width=\linewidth]{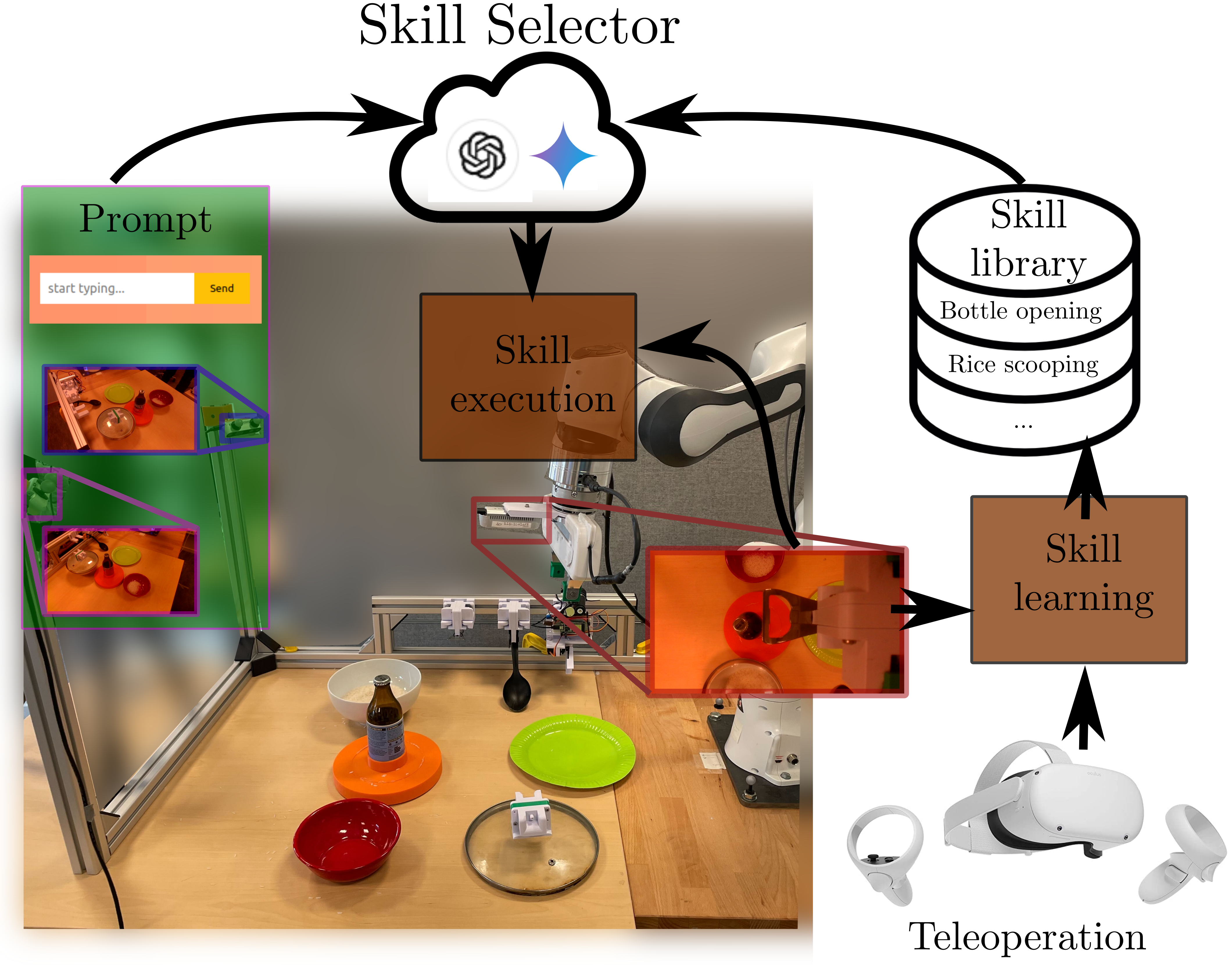}
    \caption{Conceptual overview of our Robotic Skill Learning System. The system receives the user's instructional prompt and an image of the current state. The skill selector module - realized through a foundational model - selects an appropriate skill to perform the task. If no suitable skill is available, the system asks the user to perform a number of demonstrations and train a new skill using visuomotor diffusion policies.
    }
    \label{fig:over}
\end{figure}

In this work, 
the user can continuously show and teach new skills using intuitive teleoperation. Our approach is able to receive instructions via natural language and its current observation - a procedure that most humans are now familiar with thanks to the widespread adaptation of Large Language Models~\cite{eloundou2023gpts}. The framework then consults its skill library - a repository of skills it has learned in the past - and assesses if any of the currently available skills are applicable to execute the given task. If, however, no suitable skill is available, the system will simply ask the user to provide a number of demonstrations (around $50-150$). The new skill can then be trained on external hardware and loaded onto the system when completed.
In this way, the system's capabilities can be continuously expanded by the user.
A schematic overview of our Robotic Skill Learning System (RSLS) is given in Fig. \ref{fig:over}.
The system uses a Large Language Model (LLM) and a Large Visual Language model (VLM) as foundational models to assess if any of the learned skills are applicable to fulfill the user's instructions. New skills can be easily integrated by performing demonstrations using an off-the-shelf Oculus VR controller, and training using a visuomotor diffusion policy.
We assess the RSLS in the real world
on complex skills that have to be executed sequentially such as removing a lid from a bowl before being able to access its content as well as 
extend the use of diffusion policies to contact-rich and granular material manipulation tasks. We can summarize our contributions as follows:
\begin{itemize}
    \item we present a fully functional Teaching by Demonstration framework both in simulation and the real world;
    \item we compare different large Language and Visual Language Models for skill selection;
    \item we apply diffusion-based visuomotor policy to contact-rich and granular material tasks;
    \item we conduct an extensive experimental evaluation of each skill and component of the full framework.
\end{itemize}

\section{Background \& Related Work}
In this section, we first briefly explain the required background knowledge and related work in regards to VLM and visuomotor diffusion policies. We end the section by discussing relevant work with respect to our framework.

\noindent
\textbf{Large Language/Visual Language Models as Foundational Models for Robotic Skill Selection:}

Mapping natural language descriptions to robotic actions allows for simple and relatable interaction with a robotic system. Approaches to achieve this vary, including direct conditioning of models on actions\cite{liang2023skilldiffuser}\cite{brohan2022rt}\cite{brohan2023rt}, as well as strategies that combine Large Language Models (LLMs) with visual data, creating Visual Language Models (VLMs) that can take multimodel inputs i.e. text and images or video as input. The integration of VLMs into robotics is a step towards robots that can interpret and act on natural language instructions in a visual context. Examples demonstrating the usefulness of this approach include frameworks that emulate human cognitive processes for nuanced task execution \cite{zhu2024language}, and the processing of complex, multimodal task descriptions \cite{yang2023octopus}. These advancements collectively highlight the promise of VLMs as a strategy to enhance robotic capabilities.

\noindent
\textbf{Visuomotor diffusion policies for robotic manipulation:}

Diffusion, originally popularized in the domain of generative image models, have recently emerged as a powerful tool in the field of robotics. In particular, diffusion-based policy models used for behavioral cloning have shown promising results \cite{chi2023diffusionpolicy}. At their core, diffusion models learn to gradually construct complex data distributions, starting from random noise and progressively refining this noise into structured outputs. In the context of robotics, these outputs are sequences of actions (in task space) of the robot to perform a given task.

One key benefits of diffusion-based policy models is the few number of human demonstrations needed to clone a specific behavior. \cite{chi2023diffusionpolicy} showed that only around 100 demonstrations were needed to clone complex behaviors such as flipping objects and handling liquids with over 70\% success rate while being more robust to disturbances and idle actions than other approaches such as \cite{shafiullah2022behavior}. Another benefit of using diffusion-based policy models is the lack of explicit modeling that needs to be done of the task and environment. \cite{fu2024mobile} shows how the same underlying learning method can be used to control a bi-manual mobile manipulation system to navigate indoor environments and perform kitchen tasks. In another example of it’s versatility, \cite{scheikl2023movement} shows how diffusion policy is used to handle deformable objects for the the purpose of robot-assisted surgery.

However, while those frameworks expand the use case of \cite{chi2023diffusionpolicy} significantly, which skill is executed and when is still entirely decided manually by an expert.

\noindent
\textbf{General Skills Learning Frameworks}
There have been several proposed frameworks that leverage language models in combination with learned skills to make robots capable of completing queried tasks in a scene.

\cite{liu2024moka} utilizes a visual language model to first decompose an overall goal into a series of subtasks given a scene of interactable objects. By superposing the camera view with a grid and keypoints on relevant objects, the VLM is able to communicate well-defined spatial planning for robotic manipulation. Keypoint-based navigation helps to mitigate some of the shortcomings still found in the spatial reasoning of current VLMs, but they also restrict the planning to primitive and non-complex tasks.

Another approach \cite{zhen20243dvla} leverages a 3D-LLM \cite{hong20233dllm} together with a “goal imagination” diffusion model to generate actions given a scene and a goal. The 3D-LLM is first used to condition the goal imagination process given the stated goal. When an imagined 3D scene of the goal has been generated, the 3D-LLM is used again to generate a sequence of action tokens for the manipulator to execute. While this has the potential to create more intricate action plans due to its more native 3D understanding, it is still limited to open-loop control.

The approach in \cite{chen2023playfusion} attempts to retrieve skills from unstructured play data. The play data is language labeled in hindsight which is used to condition a diffusion-based next action predictor. By introducing a quantization bottleneck in the diffusion process, this method is able to discretize the learned representations into individual finer skills. The discrete skills can then be used in new combinations to achieve novel goals, showcased on tasks such as pick and place action in a dinning table setting.

\section{Robotic Skill Learning System}
\begin{figure}
    \centering
    \includegraphics[width=\linewidth]{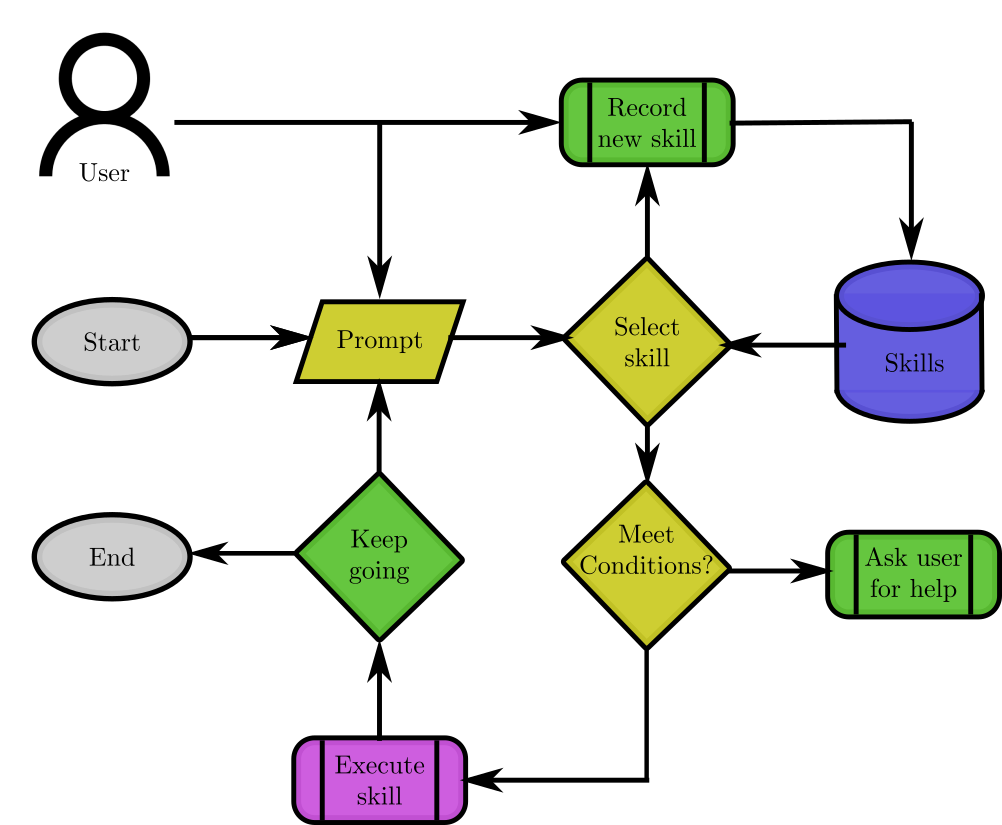}
    \caption{Flowchart overview of our RSLS method. Yellow boxes indicate the Skill Selector realized through a foundational model, green indicates the user's activity, and purple shows the visuomotor diffusion policy.}
    \label{fig:tbd_flow}
\end{figure}

In this section, we describe the Robotic Skill Learning System (RSLS) framework in detail.
Fig. \ref{fig:tbd_flow} shows an overview of our RSLS setup in the real world. It consists of a Skill Selector which is based on a 
foundational model.
Given the user's input prompt, the first step of the skill selector is to find a suitable skill in the library of skills. 
If no skill is found the system requests to be taught a new skill by the user.
In this case, the RSLS enters demonstration mode, and the user can perform repeated demonstrations of the new skill which will be, after training, added to the skill library.
If a skill is found, the skill selector checks the preconditions of the particular skill given an image of the current workspace using the multimodal aspect of the foundational model. If all preconditions are met, the skill is then sent to the Skill Execution Module. 

\subsection{Teaching a Skill}
We record the demonstrations in such a way that they are compatible with the diffusion policy training framework introduced in~\cite{chi2023diffusionpolicy}.
To facilitate teleoperating the robot we make use of the stand-alone app Quest2ROS~\cite{welle2024quest2ros} for the Oculus Quest. 
 This way, a user can easily record demonstrations for an arbitrary task using readily available hardware. We extend the range of tasks via diffusion policy to contact-rich tasks (bottle opening) as well as the handling of granular material (rice scooping task) in the real world.

\subsection{Training/Executing a Skill}
Once a number of demonstrations (depending on the skill, between $50-150$) are collected for any particular task, we train a visuomotor diffusion policy as detailed in~\cite{chi2023diffusionpolicy}. Specifically, we use a CNN-based denoising network along with separate RestNet18 encoders for each camera view. Once the skill is trained, it is added to the Skill library with a short description of the skill, as well as what preconditions have to be fulfilled, and a method to execute to use the skill, as shown in\ref{lst:task_library}.

\begin{lstlisting}[caption=An entry in the skill library,label={lst:task_library}, basicstyle=\tiny]
Skill(
    "SERVE RICE",
    "This skill serves rice from a white bowl into a red bowl",
    "The white bowl needs to contain rice. A red bowl needs to visible in the 
    workspace."
    ),
\end{lstlisting}

\subsection{Skill Selector}
The Skill Selector is realized using a large pre-trained Foundational Model.
The input is an image of the current scene as well as the user's instructional prompt. 
In the first step, the user prompt is fed into an LLM, which has access to the names and descriptions of the Skill library.
The task of the LLM is to select a suitable skill given the user's request and the names and descriptions of the skills in the skill library. If no suitable skill is identified, the system will ask the user to teach it the new required skill.

If a matching skill is found by the LLM, a second step is performed to check if all preconditions for performing the skill are met. The preconditions are sent to the VLM along with an image of the scene. As output, the VLM will have to make a ``YES" or ``NO" decision on whether the preconditions are met or not. 
If any of the preconditions are violated, the system will inform the user which can amend the situation. 
If no preconditions are violated, the execution method of the skill is called and the skill is executed.

This flow is depicted schematically in Figure \ref{fig:tbd_flow}.
Determining a suitable skill using only the LLM in the first step of the skill selector saves the query to the more computationally expensive VLM if no suitable skill is found in general. If a skill is found the VLM can be prompted more specifically to only check if the by the skill given precondition is fulfilled and the skill can be executed.

\subsection{Simulation Setup}
\begin{figure}
    \centering
    \includegraphics[width=\linewidth]{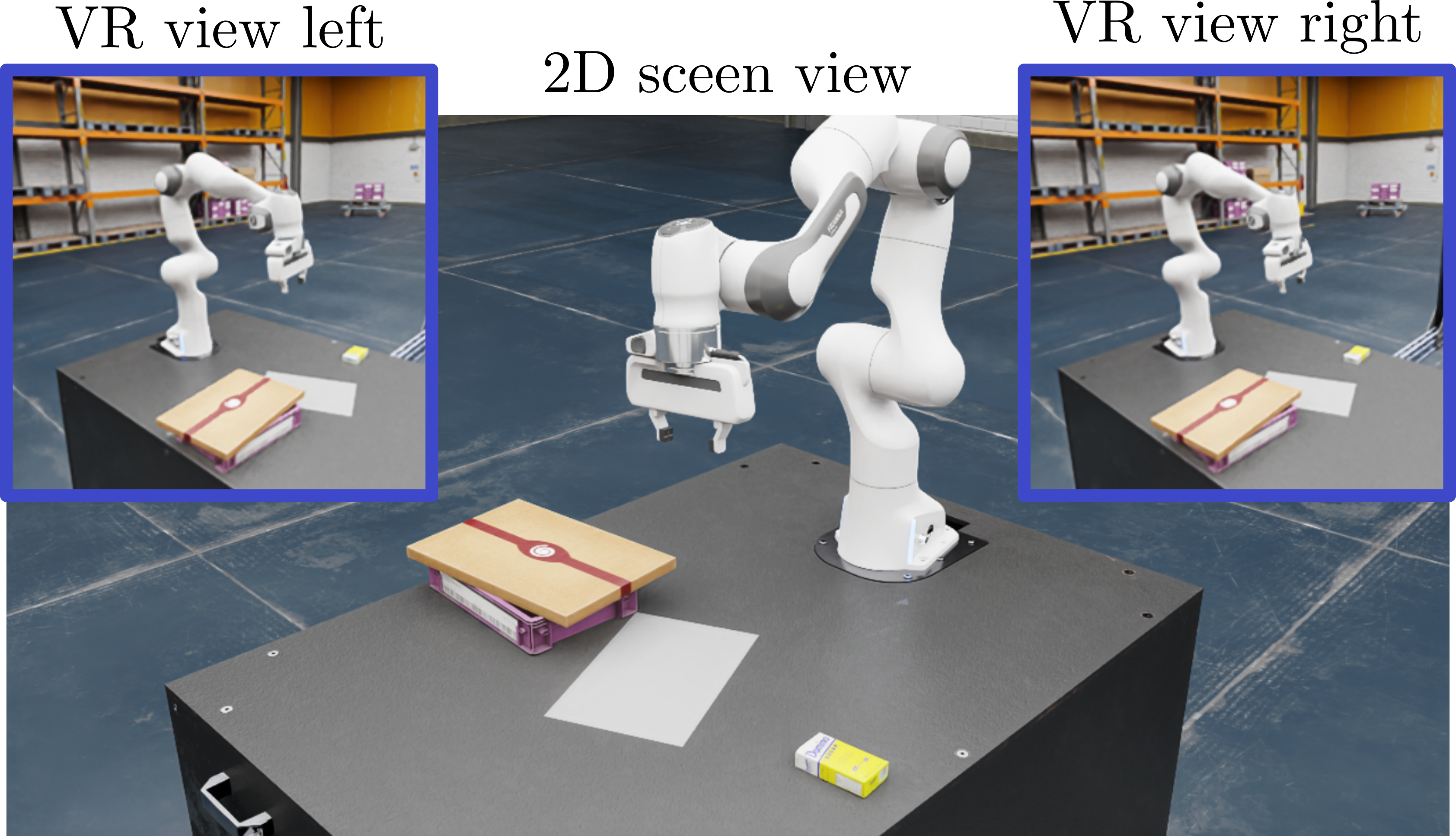}
    \caption{Setup of the simulation environment, including the VR views for the left and right eye. Note that as the user is free to traverse the virtual environment he can obtain different views than those shown on a 2D screen.
    }
    \label{fig:sim_setup}
\end{figure}

To obtain as realistic conditions for the simulation as possible we adopt
Isaac Sim with the Orbit framework~\cite{mittal2023orbit} for simulation. To make the teleoperation similar to the real world, we adapted the teleoperation framework from~\cite{welle2024quest2ros}. 
In order to successfully teleoperate a robot in 3D space, it is beneficial to display the environment in a 3D space as well; receiving teleoperation feedback from a 2D screen can result in confusion \cite{hetrick2020comparing}\cite{moletta2023virtual}.
To overcome this issue, a VR camera rig is set up in Orbit which records stereoscopic images using a virtual camera which are streamed to the Oculus headset. The cameras have an asymmetric frustum and matched parameters to the human eye to enhance the 3D experience. The movements of the Oculus in the real world in turn determine the change in position of the virtual camera in Orbit. This way the user is placed in an interactive 3D scene when performing demonstrations which make it possible to immersively and naturally teleoperate the robot in simulation using the same control scheme as in the real world. The setup can be seen in Fig \ref{fig:sim_setup} including the left and right eye images.

\subsection{Real World Setup}

\begin{figure}
    \centering
    \includegraphics[width=\linewidth]{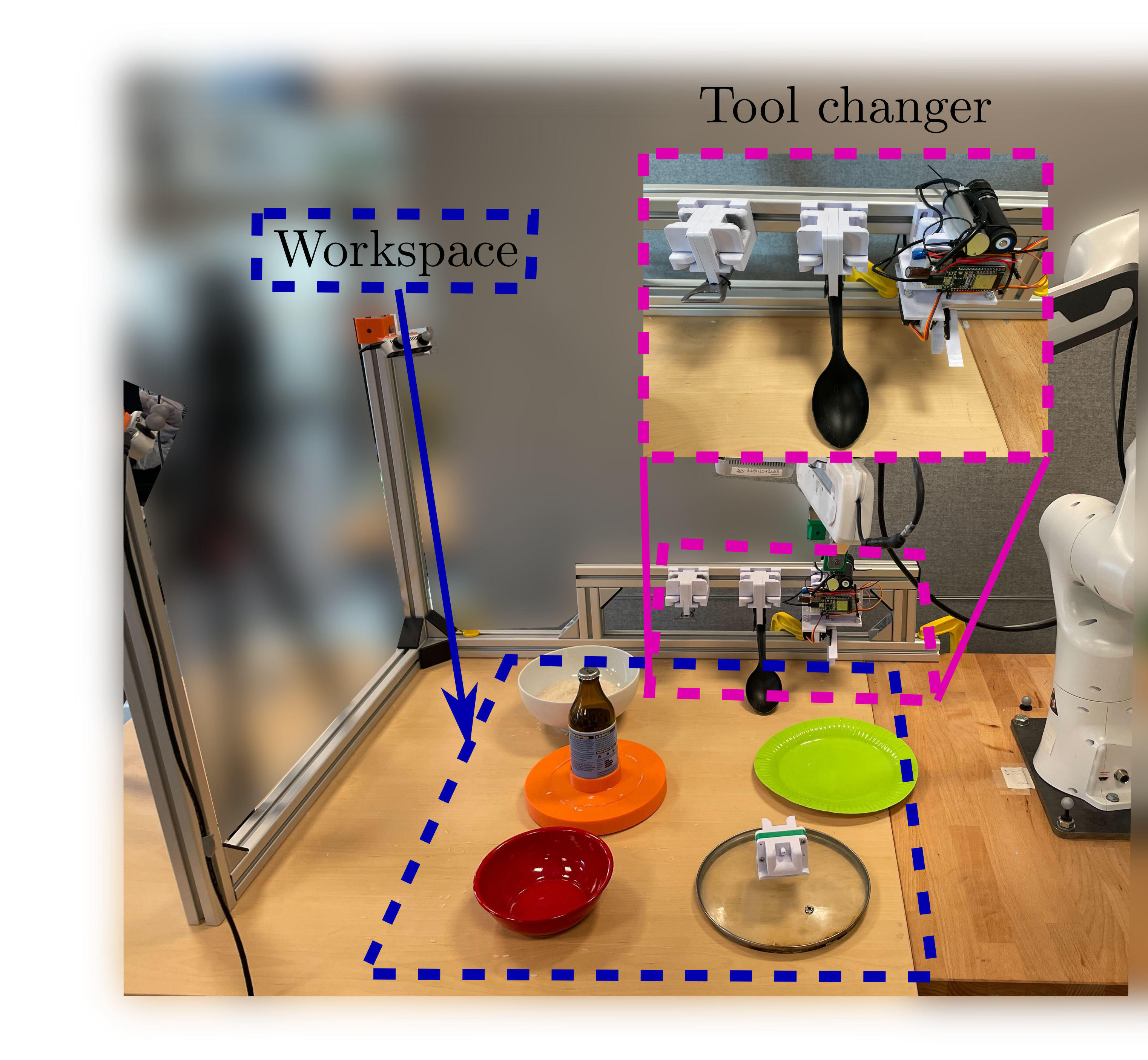}
    \caption{The  real world setting indicating the workspace (blue) and the tool changer (pink) containing a bottle opener, a serving spoon, and a custom gripper for the sausages.}
    \label{fig:setup_real}
\end{figure}
The robotic setup can be seen in Fig. \ref{fig:setup_real}, and consists out of a Franka Panda manipulator with a Realsense camera mounted on the end-effector, the robot is able to interact/manipulate with the food-related items such as a bottle, a bowl of rice with a lid and a plate of sausages, in its workspace.
As each task in the real world requires a specific tool a tool change station is also mounted in the workspace of the robot.
Furthermore, two additional real-sense cameras are mounted on the opposite side of the robot providing the skill selector with an unobstructed view of the scene.

\noindent
\textbf{Tool Changer:} the Toolchanger holds three different tools as shown in Fig. \ref{fig:setup_real}: \emph{i)} a bottle opener, \emph{ii)} a large serving spoon, and \emph{iii)} a compliant custom gripper. If a skill needs a specific tool the robot will first exchange/equip the appropriate tool for the skill. Not having any tool is also a viable option, for instance when performing a pick-and-place task such as the lid removal.

\section{Experimental Evaluation}

We first report insights from the training of new skills using experienced operators. Next, we report the performance of the trained skills individually as well as the performance of the skill selector. Finally, we validate the functionality of the full system.

\subsection{Human in the Loop}
\begin{figure}
    \centering
    \includegraphics[width=\linewidth]{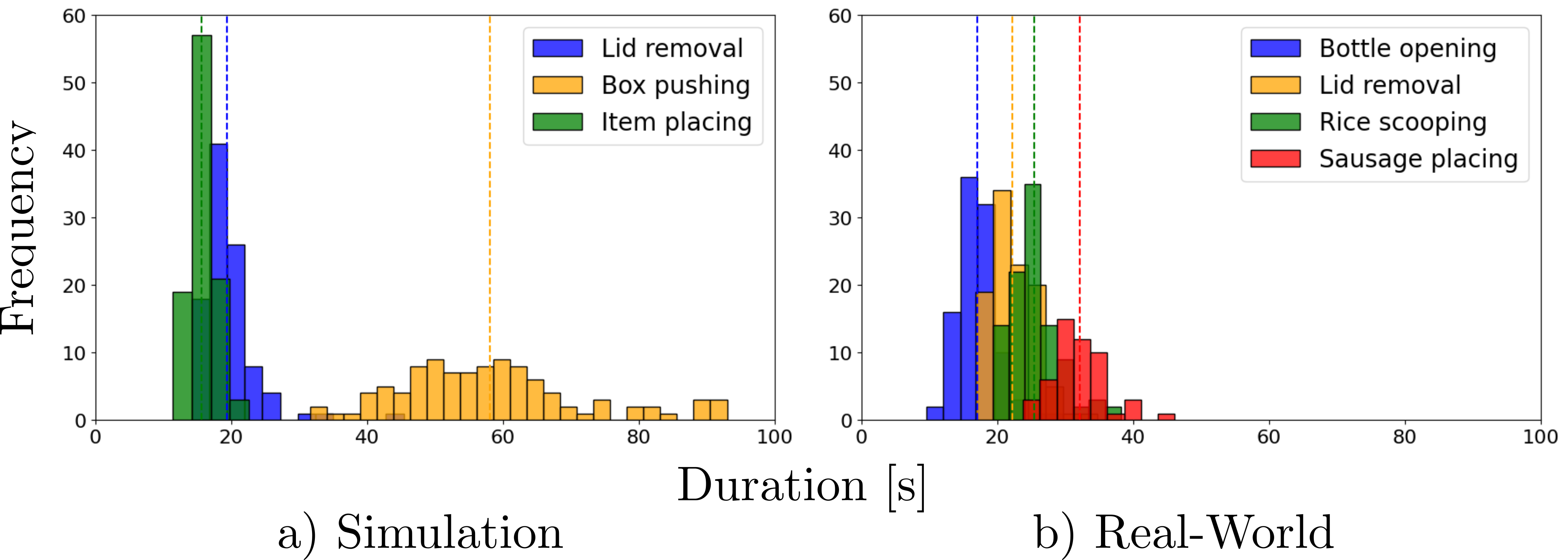}
    \caption{Demonstration time histograms for the lid removal (blue), box pushing (orange), and item placing (green) in simulation (left) as well as for the real-world tasks Bottle opening (blue), lid removal (orange), rice scooping (green), and sausage placing (red) on the right. The dashed lines indicate the mean duration of the respective task.}
    \label{fig:user_time_hist}
\end{figure}

In this work, we collected approximately 100 demonstrations per skill in simulation and the real world by two experienced operators -- authors of the paper. A total of $7$ different skills have been trained. The demonstrations are done using the stand-alone Oculus app Quest2ROS~\cite{welle2024quest2ros} in both settings. In the simulation, the functionality of the app has been extended to include head-tracking and VR rendering. Fig.~\ref{fig:user_time_hist} shows the histograms of the demonstration time for the skills learned in simulation (left) and the real world (right) respectively. The mean time for each skill is indicated by the dashed lines. 
 We can show that rather complex tasks such as opening a bottle can be demonstrated in a short amount of time i.e. around $100 \cdot 17.1s = 28.5$ min.
Overall we can see that the demonstrations follow an expected normal distribution with a mean time of $19.5$, $15.6$, and $58.1$ seconds for the cap removal, pick and placing, and crate pushing in simulation respectively. Naturally, pushing the crate to a target configuration is more complex and therefore takes a much longer time on average than the other tasks. In the real world, the tasks' mean durations are: bottle opening - $17.1$s, lid removal - $22.3$s, rice scooping - $25.5$s, and sausage placing - $32.2$s. The sausage placing only received $50$ demonstrations, as a single demonstration includes placing three sausages into the red bowl.

Observations are collected at $0.1$s intervals during demonstrations. The simulation is played at half speed to ensure a sufficiently low computational load and ensure physics interactions can be computed correctly.

\subsection{Simulation Setting}
As shown in Fig. \ref{fig:sim_setup}, the simulation setting consists of three different tasks. \emph{i)} Cap removal -  remove the cap from the box,
\emph{ii)} crate pushing - push the crate onto the designated target area, and \emph{iii)} picking and placing - picking up a box of sugar and putting it into the box.
We report the results of all three tasks in Table \ref{tab:simulation_experiments}, including the number of demonstrations used for training and the success criteria. Furthermore, the videos of all demonstrations and evaluations can be found on the project website\footref{fn:website}.

\noindent\textbf{Implementation details: }The data collected for the training of the policies is similar to the real world. With a frequency of $10$Hz, the following observations are logged: A 4x4 transformation matrix from the robot base to the end-effector, A 240x320 image from an end-effector camera, and the actions recorded from the oculus controller, which is a 6D velocity vector in end-effector space 
along with a scalar value that acts as the gripper activation signal.
After each demonstration, the scene can be easily reset by the user, which randomly initializes the pose of the purple crate and its cap, the gray plane, as well as the box of sugar.

When playing the policies, a parallel thread is started to perform inference at a frequency of 4Hz. The action horizon is then updated as soon as the inference has finished.

\begin{table*}[t]
  \centering
  \begin{tabular*}{\linewidth}{||p{0.1\linewidth}|c|p{0.4051\linewidth}|c|c|p{0.15\linewidth}||}
    \hline
    
    Name & $N_{Demo}$ & Success criteria & Time limit & Success rate & Notes \\

    \hline
    \hline
    
    Cap removal & 100 & The geometric center of the lid is more than it's length away from the geometric center of the crate and the crate is on top of the table. &$20$s & 83/100 & \\
    
    Crate pushing \newline (single view)& 100 & The Intersection over Union (IoU) of the crate and goal marking is larger than 0.8. & $60$s & 20/100 & Average IuO: $39.0\%$\\

    Crate pushing \newline (multi view)& 100 & The Intersection over Union (IoU) of the crate and goal marking is larger than 0.8. & $60$s & 74/100 & EE, front $\And$ side view. Average IuO: $83.4\%$\\
    
    Pick and Placing & 100 & The geometric center of the sugar box is inside the convex hull of the crate. & $20$s & 96/100 & \\
    
    \hline
    
  \end{tabular*}
  \caption{Overview of all simulated policy experiments.}
  \label{tab:simulation_experiments}
\end{table*}

All trained policies are evaluated by $100$ trial runs. The success criteria and rates are shown in table \ref{tab:simulation_experiments}. The results are discussed for each experiment below.

\noindent
\textbf{Cap removal:} This task reaches a success rate of $83\%$, the only failure cases occur when the robot did not only push the cap but the crate as well.

\noindent
\textbf{Crate pushing:} When training and running this task on only the end-effector view the performance is very poor - only $20\%$ reached a IoU $>0.8$. This is expected, as the end-effector view barely contains any information about the position of the crate relative to the goal. 
When adding an additional front and side view the policy receives additional information containing more spatial cues that can inform the actions leading to a success rate of $74\%$. Still, the policy can end up in cases where the diffused actions are very small resulting in the robot staying in place and not pushing the box anymore. 
Another failure case is created due to errors in the physics simulation. Occasionally, the gripper would clip into the crate during pushing and the two objects can not move independently anymore.

\noindent
\textbf{Pick and placing:} With $96\%$ this is the highest scoring task, the failures happened because of the sugar box being dropped on the edge of the crate.

\subsection{Real World Tasks}
As shown in Fig. \ref{fig:setup_real} the real-world setting consists of four different skills in a food serving setting, and each skill requires a dedicated tool. The pose of the objects of the table is varied between all experiments
. The individual skills are: \emph{i)} bottle opening - using the opener to remove the bottle cap from the bottle,  \emph{ii)} lid removal - removing the lid on top of the white rice bowl using the gripper, \emph{iii)} rice scooping - transferring rice from the white bowl into the red bowl using the serving spoon, and \emph{iv)} sausage placing - placing sausages from the green plate into the red bowl using the custom gripper. 

\textbf{Implementation details:} The policy model was configured to operate at a control frequency of $10$ Hz, using the observations from $2$ previous iterations and predicting $14$ timesteps into the future. The first $8$ of these timesteps were then executed on the robot before generating $14$ new actions and repeating the process. 
All skills were conditioned on the observations from the end-effector camera.
The camera feed was first resized to $240x320$ before being fed into the encoder and the overall policy model. The robot state represented as a 
$4x4$ transformation matrix from the base to the end effector
was also used as observations. Each skill was trained for $600$ epochs, taking approximately $10$ h on a NVIDIA GeForce RTX $4090$.

The performance of all individual skills is shown in Table \ref{tab:real_world_experiments}.
All execution videos of all experiments are available on the project website\footref{fn:website}.

\noindent
\textbf{Bottle opening:} Constitutes the most challenging task reaching $60\%$. The main failure case is when the robot misses the bottle when approaching it from above. When the bottle opener gets in a position relative to the bottle that is out of the dataset distribution.
In one such failure case, the robot was able to recover but just outside the time window for successful task completion. Whenever the robot was able to latch onto the cap, a successful task completion always followed swiftly.

\noindent
\textbf{Lid removal:} Succeded $90\%$, in the single failure case, the grippers was not successful in centering itself when approaching the lid from above. This resulted in one of the gripper fingers getting stuck on the handle and the robot was unable to recover.

\noindent
\textbf{Rice scooping:} Reached $90\%$ as well. This policy was given $90$ seconds 
due to its complex nature. In the single failure case of the nine tries, the robot was able to scoop up a sufficient amount of rice and transfer it over to the red bowl. However, it never tilted the spoon to let the rice into the bowl.

\noindent
\textbf{Sausage placing:} Performed successfully in $90\%$ of the cases and the single failure case occurred when the gripper dropped one of the sausages over the green plate and it landed in a position it was not able to be picked up from.

\begin{table*}[t]
  \centering
  \begin{tabular*}{\linewidth}{||p{0.11\linewidth}|c|p{0.57\linewidth}|c|c||}
    \hline
    
    Name & $N_{Demo}$ & Success criteria & Time limit & Success rate \\
    
    \hline
    \hline
    
    Rice scooping & 101 & At least $5$ grams of rice has been moved from the white bowl into the red bowl. & $90$s & 9/10 \\
    
    Bottle opening & 100 & The bottle cap is fully removed from the bottle. & $60$s & 6/10 \\
    
    Lid removal & 100 & The lid is placed onto the table. & $60$s & 9/10 \\
    
    Sausage placing & 50 & All three sausages have been moved from the green plate into the red bowl. & $60$s & 9/10 \\
    
    \hline
    
  \end{tabular*}
  \caption{Overview of all real-world policy experiments.}
  \label{tab:real_world_experiments}
\end{table*}

\subsection{Skill Selector}
The skill selector's job is to select an adequate skill given a user prompt or
request a new skill, once a skill is found an image of the scene is used to determine if the preconditions for the skill are fulfilled.
We compare two state-of-the-art foundational models, GPT-4 \cite{openai2024gpt4} and Gemini \cite{geminiteam2023gemini} (accessed 15/03/2024).

Evaluating foundational models is notoriously difficult and still an open research direction \cite{LLMBenchmarkSurvey}. For that reason, the models are evaluated on our specific use cases. One evaluation is performed for the skill matching step and one for the precondition validation step.

The skill matching evaluation is set up as follows: For each $16$ combinatorial variation of the four skills in the skill library (including no skill at all), each four skills are requested using two variations for the user prompt. The experiment is repeated five times resulting in a total of $640$ prompts and responses to evaluate. The skills in the library and the user prompt are fed into the foundational model using the template in Listing \ref{lst:llm_prompt}. The descriptions and preconditions of the skills are stated in Listing \ref{lst:skill_library}. The user prompts and corresponding skills are stated in Listing \ref{lst:llm_user_inputs}.
The response is considered correct if it includes the skill name the user asked for and this skill is indeed in the library. 
If the requested skill is not in the library, the response is considered correct if it
requests a new skill.
The results are shown in Table \ref{tab:ss_performance}.

\begin{lstlisting}[caption=Prompt template to the LLM the skills and user input get injected to the placeholders,label={lst:llm_prompt},breaklines=true, basicstyle=\tiny]
You are an expert skill selector that has to match skills that are given to a 
user's request. If none of the skills given to you are fulfilling the users 
request, answer with "NEW SKILL".

Your skills are:
[[[SKILL NAMES AND DESCRIPTIONS]]]

User request:
[[[PROMPT]]]

Structure your answer in this format:
[reasoning without metioning the names of skills]
[Skill Name]
\end{lstlisting}

\begin{lstlisting}[caption=Description and preconditions of the four skills in the skill library, label={lst:skill_library},breaklines=true, basicstyle=\tiny]
Skill(
    "SERVE RICE",
    "This skill serves rice from a white bowl into a red bowl",
    "The white bowl needs to contain rice. A red bowl needs to visible in the workspace."
    ),
Skill(
    "OPEN BEER",
    "This action opens the beer bottle by removing the metal cap",
    "The bottle needs to be closed with a metal cap"
    ),
Skill(
    "SERVE SAUSAGE",
    "This skill picks up one or more sausages from a green plate and puts them into a red bowl",
    "A green plate with sausages on them needs to be visible in the workspace. A red bowl needs to be visible in the workspace which could contain something already.",
    ),
Skill(
    "REMOVE LID",
    "This skill removes the glass pan cover from the white bowl of rice.",
    "A glass pan cover has to be present and not on the table.",
    )
\end{lstlisting}

\begin{lstlisting}[caption=Mock user requests used for the evaluation of skill matching.,breaklines=true, label={lst:llm_user_inputs}, basicstyle=\tiny]
Rice scooping: ["Serve the rice please.", "I want rice!"]
Bottle opening: ["Open the bottle!", "I would like something to drink please."]
Sausage placing: ["Give me some meat!", "Please move the sausages from the green plate to the red bowl"]
Lid removal: ["Please remove the lid from the rice.", "Uncover the rice!"]
\end{lstlisting}

In order to evaluate the VLM, we collected $10$ images of the scene for each permutation of skills that are able to be performed.
In total $110$ pictures were collected. For each picture, the preconditions are validated for all four skills, leading to $440$ prompts and responses to evaluate.

The preconditions shown in Listing \ref{lst:skill_library} are fed into the VLM using the template in Listing \ref{lst:vlm_prompt} along with an image of the scene. A response is considered correct if a "YES" or "NO" is retrieved from the last line of the response which matches the ground truth of whether all preconditions are met.

The experiment is repeated three times, using images from a camera on the left side of the scene, images from the right side, and once using both images. The results are shown in Table \ref{tab:ss_performance}.
\begin{lstlisting}[caption=Prompt template to the VLM the precondtion gets injected into the placholder,label={lst:vlm_prompt},basicstyle=\tiny]
Please check if the following conditions are met in the image:
[[[PRECONDITIONS]]]

Answer format for each precondition:
[Short Reasoning]
[YES/NO]

End the response with a definitive answer (YES/NO) on whether ALL conditions are 
met on a new line.
\end{lstlisting}

\begin{table}[h!]
\centering
\begin{tabular}{||c | c | c|c|c|c||} 
 \hline
 Model & LLM  & VLM-l & VLM-r & VLM-lr & LLM$\cdot$VLM-r \\ 
 \hline\hline
 GPT-4          & $\bold{96.3}$\% & $71.1$\% & $\bold{77.5}$\% & $71.9$\% & $\bold{74.6}$\% \\ 
 Gemini          & $93.0$\%& $69.1\%$ & $75.7\%$  & $65.0\%$ & $70.4\%$ \\
 \hline
\end{tabular}
\caption{Comparison of GPT-4 and Gemini as foundational models for skill selection. Best results in bold.}
\label{tab:ss_performance}
\end{table}

All prompts and responses for both the LLM and VLM evaluation can be found on the website\footref{fn:website}. From Table \ref{tab:ss_performance}, it can be seen that the LLM variant of both foundational models is able to achieve a high result matching the user request to the skill library at hand. Precondition checking seems to be slightly more difficult. Having a good camera angle seems to matter, as well as that more cameras do not necessarily result into better performance, 
which might be attributed to the fact that more information can lead to confusing answers of the VLM.
A multiplicative result is shown taking the best result from the LLM evaluation together with the best result of the VLM evaluation. This is a good indicator for the success rate of a request to the robot in the full system.
GPT-4 performed slightly better than Gemini and was subsequently used in the full systems.

\begin{figure}
    \centering
    \includegraphics[width=\linewidth]{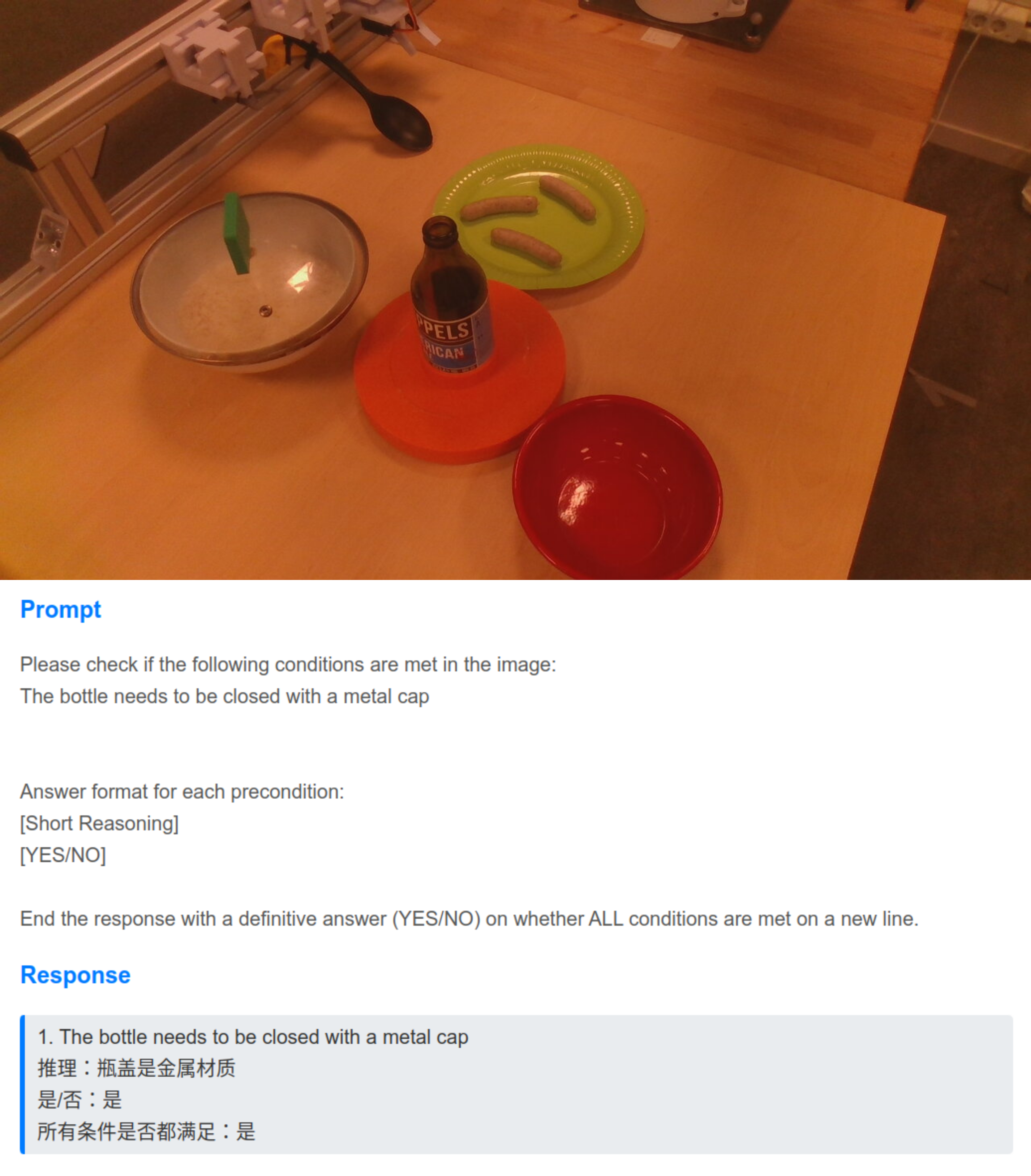}
    \caption{Peculiar failure case of the model when given an image and a prompt, the response is largely in Chinese.}
    \label{fig:fm_f_case}
\end{figure}

\noindent
\textbf{Peculiar failure case:} Fig. \ref{fig:fm_f_case} shows a particular peculiar failure case of the foundational model when prompted to check the preconditions for the bottle opening, the model responded in Chinese characters instead with the prompting and location language of English. This kind of unexpected failure mode highlights the need for further studies and additional output checks when deploying these models in combination with other technology.

\subsection{Validation of SSLE Framework}
\begin{figure}
    \centering
    \includegraphics[width=\linewidth]{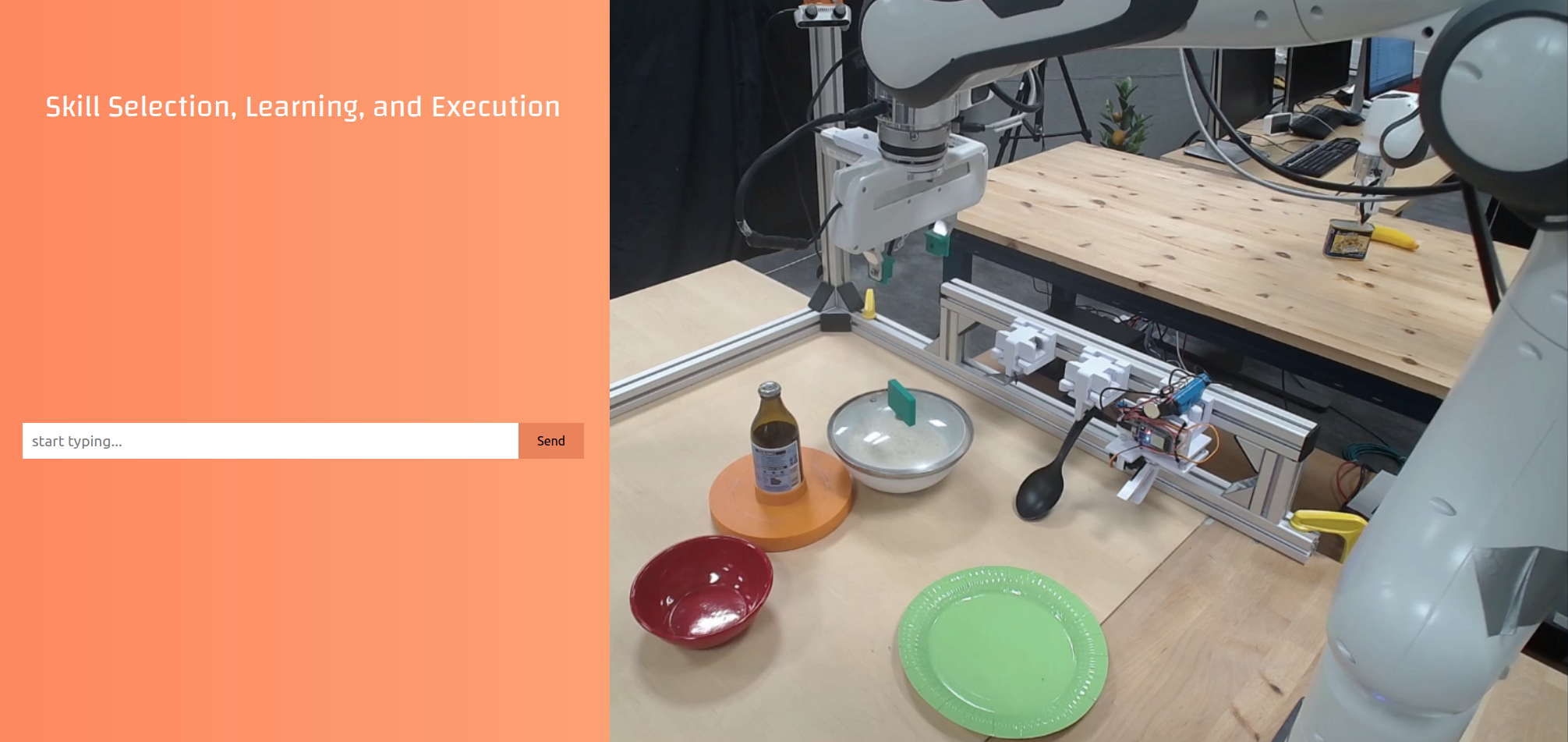}
    \caption{The setup used for validation, with a prompting window (left) and the workspace (right)}
    \label{fig:Validation_Setup}
\end{figure}
As a user study lies outside of the scope of this work we validate the full framework by playing out an interaction with a typical user. The validation aims to show the human-in-the-loop interactions when new skills are required and when skills can not be performed due to precondition validation. We recommend to watch the validation video on the website\footref{fn:website}.

The interaction is as follows: the user arrives at the scene as shown in Fig. \ref{fig:Validation_Setup}. For now, the tasks in the skill library are "Bottle opening", "Rice scooping" and "Sausage placing". The user starts by saying they are thirsty and would like a refreshment. The framework finds a suitable skill (Bottle opening), validates the preconditions, and executes the skill. 
Next, the user asks to remove the pan cover from the rice. The system replies that no suitable skill is available and requests to teach a "new" lid removal skill. At this point, the skill library is reloaded with the additional "Lid removal" skill to simulate the act of having shown and learned a new skill. When the user repeats the prompt in regards to removing the pan cover, this action is now executed.
After successfully removing the pan cover, the user asks to move the rice to the red bowl. The system matches the rice scooping skill, validates the preconditions, and executes the action. 
Finally, the user prompts the system to provide some sausages. Since there are no sausages on the plate yet, the system returns that while it has the skill to do so, it can not perform it since the preconditions are not met. After placing some sausages on the plate and re-prompting, the system executes the sausage placing skill.

\section{conclusion}
In this work, we presented a 
Robotic Skill Learning System that builds upon 
diffusion policies and foundational models. Our system is able to learn novel tasks via diffusion policies using approximately $100$ demonstrations per task given by the user. We compared two state-of-the-art foundational LLMs/VLMs in their role to select a known skill from a skill library or ask for a new skill as well as their capability to check preconditions and determine if the skill should be executed or not. We extensively evaluated the individual skills, and parts of the system with all detailed results public on the project website\footref{fn:website}, and validated the whole framework as shown in the supplementary video.


\balance
\printbibliography

@article{shafiullah2022behavior,
  title={Behavior transformers: Cloning $ k $ modes with one stone},
  author={Shafiullah, Nur Muhammad and Cui, Zichen and Altanzaya, Ariuntuya Arty and Pinto, Lerrel},
  journal={Advances in neural information processing systems},
  volume={35},
  pages={22955--22968},
  year={2022}
}

@article{mittal2023orbit,
  title={Orbit: A unified simulation framework for interactive robot learning environments},
  author={Mittal, Mayank and Yu, Calvin and Yu, Qinxi and Liu, Jingzhou and Rudin, Nikita and Hoeller, David and Yuan, Jia Lin and Singh, Ritvik and Guo, Yunrong and Mazhar, Hammad and others},
  journal={IEEE Robotics and Automation Letters},
  year={2023},
  publisher={IEEE}
}

@inproceedings{welle2024quest2ros,
  title={Quest2ROS: An App to Facilitate Teleoperating Robots},
  author={Welle, Michael C and Ingelhag, Nils and Lippi, Martina and Wozniak, Maciej and Gasparri, Andrea and Kragic, Danica},
  booktitle={7th International Workshop on Virtual, Augmented, and Mixed-Reality for Human-Robot Interactions},
  year={2024}
}

@inproceedings{hetrick2020comparing,
  title={Comparing virtual reality interfaces for the teleoperation of robots},
  author={Hetrick, Rebecca and Amerson, Nicholas and Kim, Boyoung and Rosen, Eric and de Visser, Ewart J and Phillips, Elizabeth},
  booktitle={2020 Systems and Information Engineering Design Symposium (SIEDS)},
  pages={1--7},
  year={2020},
  organization={IEEE}
}

@inproceedings{moletta2023virtual,
  title={A virtual reality framework for human-robot collaboration in cloth folding},
  author={Moletta, Marco and Wozniak, Maciej K and Welle, Michael C and Kragic, Danica},
  booktitle={2023 IEEE-RAS 22nd International Conference on Humanoid Robots (Humanoids)},
  pages={1--7},
  year={2023},
  organization={IEEE}
}

@article{LLMBenchmarkSurvey,
author = {Chang, Yupeng and Wang, Xu and Wang, Jindong and Wu, Yuan and Yang, Linyi and Zhu, Kaijie and Chen, Hao and Yi, Xiaoyuan and Wang, Cunxiang and Wang, Yidong and Ye, Wei and Zhang, Yue and Chang, Yi and Yu, Philip S. and Yang, Qiang and Xie, Xing},
title = {A Survey on Evaluation of Large Language Models},
year = {2024},
publisher = {Association for Computing Machinery},
address = {New York, NY, USA},
issn = {2157-6904},
url = {https://doi.org/10.1145/3641289},
doi = {10.1145/3641289},
journal = {ACM Trans. Intell. Syst. Technol.},
month = {jan}
}

@misc{geminiteam2023gemini,
      title={Gemini: A Family of Highly Capable Multimodal Models}, 
      author={Gemini Team and Rohan Anil and Sebastian Borgeaud and Yonghui Wu and Jean-Baptiste Alayrac and Jiahui Yu and Radu Soricut and Johan Schalkwyk and Andrew M. Dai and Anja Hauth and Katie Millican and David Silver and Slav Petrov and Melvin Johnson and Ioannis Antonoglou and Julian Schrittwieser and Amelia Glaese and Jilin Chen and Emily Pitler and Timothy Lillicrap and Angeliki Lazaridou and Orhan Firat and James Molloy and Michael Isard and Paul R. Barham and Tom Hennigan and Benjamin Lee and Fabio Viola and Malcolm Reynolds and Yuanzhong Xu and Ryan Doherty and Eli Collins and Clemens Meyer and Eliza Rutherford and Erica Moreira and Kareem Ayoub and Megha Goel and George Tucker and Enrique Piqueras and Maxim Krikun and Iain Barr and Nikolay Savinov and Ivo Danihelka and Becca Roelofs and Anaïs White and Anders Andreassen and Tamara von Glehn and Lakshman Yagati and Mehran Kazemi and Lucas Gonzalez and Misha Khalman and Jakub Sygnowski and Alexandre Frechette and Charlotte Smith and Laura Culp and Lev Proleev and Yi Luan and Xi Chen and James Lottes and Nathan Schucher and Federico Lebron and Alban Rrustemi and Natalie Clay and Phil Crone and Tomas Kocisky and Jeffrey Zhao and Bartek Perz and Dian Yu and Heidi Howard and Adam Bloniarz and Jack W. Rae and Han Lu and Laurent Sifre and Marcello Maggioni and Fred Alcober and Dan Garrette and Megan Barnes and Shantanu Thakoor and Jacob Austin and Gabriel Barth-Maron and William Wong and Rishabh Joshi and Rahma Chaabouni and Deeni Fatiha and Arun Ahuja and Ruibo Liu and Yunxuan Li and Sarah Cogan and Jeremy Chen and Chao Jia and Chenjie Gu and Qiao Zhang and Jordan Grimstad and Ale Jakse Hartman and Martin Chadwick and Gaurav Singh Tomar and Xavier Garcia and Evan Senter and Emanuel Taropa and Thanumalayan Sankaranarayana Pillai and Jacob Devlin and Michael Laskin and Diego de Las Casas and Dasha Valter and Connie Tao and Lorenzo Blanco and Adrià Puigdomènech Badia and David Reitter and Mianna Chen and Jenny Brennan and Clara Rivera and Sergey Brin and Shariq Iqbal and Gabriela Surita and Jane Labanowski and Abhi Rao and Stephanie Winkler and Emilio Parisotto and Yiming Gu and Kate Olszewska and Yujing Zhang and Ravi Addanki and Antoine Miech and Annie Louis and Laurent El Shafey and Denis Teplyashin and Geoff Brown and Elliot Catt and Nithya Attaluri and Jan Balaguer and Jackie Xiang and Pidong Wang and Zoe Ashwood and Anton Briukhov and Albert Webson and Sanjay Ganapathy and Smit Sanghavi and Ajay Kannan and Ming-Wei Chang and Axel Stjerngren and Josip Djolonga and Yuting Sun and Ankur Bapna and Matthew Aitchison and Pedram Pejman and Henryk Michalewski and Tianhe Yu and Cindy Wang and Juliette Love and Junwhan Ahn and Dawn Bloxwich and Kehang Han and Peter Humphreys and Thibault Sellam and James Bradbury and Varun Godbole and Sina Samangooei and Bogdan Damoc and Alex Kaskasoli and Sébastien M. R. Arnold and Vijay Vasudevan and Shubham Agrawal and Jason Riesa and Dmitry Lepikhin and Richard Tanburn and Srivatsan Srinivasan and Hyeontaek Lim and Sarah Hodkinson and Pranav Shyam and Johan Ferret and Steven Hand and Ankush Garg and Tom Le Paine and Jian Li and Yujia Li and Minh Giang and Alexander Neitz and Zaheer Abbas and Sarah York and Machel Reid and Elizabeth Cole and Aakanksha Chowdhery and Dipanjan Das and Dominika Rogozińska and Vitaly Nikolaev and Pablo Sprechmann and Zachary Nado and Lukas Zilka and Flavien Prost and Luheng He and Marianne Monteiro and Gaurav Mishra and Chris Welty and Josh Newlan and Dawei Jia and Miltiadis Allamanis and Clara Huiyi Hu and Raoul de Liedekerke and Justin Gilmer and Carl Saroufim and Shruti Rijhwani and Shaobo Hou and Disha Shrivastava and Anirudh Baddepudi and Alex Goldin and Adnan Ozturel and Albin Cassirer and Yunhan Xu and Daniel Sohn and Devendra Sachan and Reinald Kim Amplayo and Craig Swanson and Dessie Petrova and Shashi Narayan and Arthur Guez and Siddhartha Brahma and Jessica Landon and Miteyan Patel and Ruizhe Zhao and Kevin Villela and Luyu Wang and Wenhao Jia and Matthew Rahtz and Mai Giménez and Legg Yeung and Hanzhao Lin and James Keeling and Petko Georgiev and Diana Mincu and Boxi Wu and Salem Haykal and Rachel Saputro and Kiran Vodrahalli and James Qin and Zeynep Cankara and Abhanshu Sharma and Nick Fernando and Will Hawkins and Behnam Neyshabur and Solomon Kim and Adrian Hutter and Priyanka Agrawal and Alex Castro-Ros and George van den Driessche and Tao Wang and Fan Yang and Shuo-yiin Chang and Paul Komarek and Ross McIlroy and Mario Lučić and Guodong Zhang and Wael Farhan and Michael Sharman and Paul Natsev and Paul Michel and Yong Cheng and Yamini Bansal and Siyuan Qiao and Kris Cao and Siamak Shakeri and Christina Butterfield and Justin Chung and Paul Kishan Rubenstein and Shivani Agrawal and Arthur Mensch and Kedar Soparkar and Karel Lenc and Timothy Chung and Aedan Pope and Loren Maggiore and Jackie Kay and Priya Jhakra and Shibo Wang and Joshua Maynez and Mary Phuong and Taylor Tobin and Andrea Tacchetti and Maja Trebacz and Kevin Robinson and Yash Katariya and Sebastian Riedel and Paige Bailey and Kefan Xiao and Nimesh Ghelani and Lora Aroyo and Ambrose Slone and Neil Houlsby and Xuehan Xiong and Zhen Yang and Elena Gribovskaya and Jonas Adler and Mateo Wirth and Lisa Lee and Music Li and Thais Kagohara and Jay Pavagadhi and Sophie Bridgers and Anna Bortsova and Sanjay Ghemawat and Zafarali Ahmed and Tianqi Liu and Richard Powell and Vijay Bolina and Mariko Iinuma and Polina Zablotskaia and James Besley and Da-Woon Chung and Timothy Dozat and Ramona Comanescu and Xiance Si and Jeremy Greer and Guolong Su and Martin Polacek and Raphaël Lopez Kaufman and Simon Tokumine and Hexiang Hu and Elena Buchatskaya and Yingjie Miao and Mohamed Elhawaty and Aditya Siddhant and Nenad Tomasev and Jinwei Xing and Christina Greer and Helen Miller and Shereen Ashraf and Aurko Roy and Zizhao Zhang and Ada Ma and Angelos Filos and Milos Besta and Rory Blevins and Ted Klimenko and Chih-Kuan Yeh and Soravit Changpinyo and Jiaqi Mu and Oscar Chang and Mantas Pajarskas and Carrie Muir and Vered Cohen and Charline Le Lan and Krishna Haridasan and Amit Marathe and Steven Hansen and Sholto Douglas and Rajkumar Samuel and Mingqiu Wang and Sophia Austin and Chang Lan and Jiepu Jiang and Justin Chiu and Jaime Alonso Lorenzo and Lars Lowe Sjösund and Sébastien Cevey and Zach Gleicher and Thi Avrahami and Anudhyan Boral and Hansa Srinivasan and Vittorio Selo and Rhys May and Konstantinos Aisopos and Léonard Hussenot and Livio Baldini Soares and Kate Baumli and Michael B. Chang and Adrià Recasens and Ben Caine and Alexander Pritzel and Filip Pavetic and Fabio Pardo and Anita Gergely and Justin Frye and Vinay Ramasesh and Dan Horgan and Kartikeya Badola and Nora Kassner and Subhrajit Roy and Ethan Dyer and Víctor Campos and Alex Tomala and Yunhao Tang and Dalia El Badawy and Elspeth White and Basil Mustafa and Oran Lang and Abhishek Jindal and Sharad Vikram and Zhitao Gong and Sergi Caelles and Ross Hemsley and Gregory Thornton and Fangxiaoyu Feng and Wojciech Stokowiec and Ce Zheng and Phoebe Thacker and Çağlar Ünlü and Zhishuai Zhang and Mohammad Saleh and James Svensson and Max Bileschi and Piyush Patil and Ankesh Anand and Roman Ring and Katerina Tsihlas and Arpi Vezer and Marco Selvi and Toby Shevlane and Mikel Rodriguez and Tom Kwiatkowski and Samira Daruki and Keran Rong and Allan Dafoe and Nicholas FitzGerald and Keren Gu-Lemberg and Mina Khan and Lisa Anne Hendricks and Marie Pellat and Vladimir Feinberg and James Cobon-Kerr and Tara Sainath and Maribeth Rauh and Sayed Hadi Hashemi and Richard Ives and Yana Hasson and YaGuang Li and Eric Noland and Yuan Cao and Nathan Byrd and Le Hou and Qingze Wang and Thibault Sottiaux and Michela Paganini and Jean-Baptiste Lespiau and Alexandre Moufarek and Samer Hassan and Kaushik Shivakumar and Joost van Amersfoort and Amol Mandhane and Pratik Joshi and Anirudh Goyal and Matthew Tung and Andrew Brock and Hannah Sheahan and Vedant Misra and Cheng Li and Nemanja Rakićević and Mostafa Dehghani and Fangyu Liu and Sid Mittal and Junhyuk Oh and Seb Noury and Eren Sezener and Fantine Huot and Matthew Lamm and Nicola De Cao and Charlie Chen and Gamaleldin Elsayed and Ed Chi and Mahdis Mahdieh and Ian Tenney and Nan Hua and Ivan Petrychenko and Patrick Kane and Dylan Scandinaro and Rishub Jain and Jonathan Uesato and Romina Datta and Adam Sadovsky and Oskar Bunyan and Dominik Rabiej and Shimu Wu and John Zhang and Gautam Vasudevan and Edouard Leurent and Mahmoud Alnahlawi and Ionut Georgescu and Nan Wei and Ivy Zheng and Betty Chan and Pam G Rabinovitch and Piotr Stanczyk and Ye Zhang and David Steiner and Subhajit Naskar and Michael Azzam and Matthew Johnson and Adam Paszke and Chung-Cheng Chiu and Jaume Sanchez Elias and Afroz Mohiuddin and Faizan Muhammad and Jin Miao and Andrew Lee and Nino Vieillard and Sahitya Potluri and Jane Park and Elnaz Davoodi and Jiageng Zhang and Jeff Stanway and Drew Garmon and Abhijit Karmarkar and Zhe Dong and Jong Lee and Aviral Kumar and Luowei Zhou and Jonathan Evens and William Isaac and Zhe Chen and Johnson Jia and Anselm Levskaya and Zhenkai Zhu and Chris Gorgolewski and Peter Grabowski and Yu Mao and Alberto Magni and Kaisheng Yao and Javier Snaider and Norman Casagrande and Paul Suganthan and Evan Palmer and Geoffrey Irving and Edward Loper and Manaal Faruqui and Isha Arkatkar and Nanxin Chen and Izhak Shafran and Michael Fink and Alfonso Castaño and Irene Giannoumis and Wooyeol Kim and Mikołaj Rybiński and Ashwin Sreevatsa and Jennifer Prendki and David Soergel and Adrian Goedeckemeyer and Willi Gierke and Mohsen Jafari and Meenu Gaba and Jeremy Wiesner and Diana Gage Wright and Yawen Wei and Harsha Vashisht and Yana Kulizhskaya and Jay Hoover and Maigo Le and Lu Li and Chimezie Iwuanyanwu and Lu Liu and Kevin Ramirez and Andrey Khorlin and Albert Cui and Tian LIN and Marin Georgiev and Marcus Wu and Ricardo Aguilar and Keith Pallo and Abhishek Chakladar and Alena Repina and Xihui Wu and Tom van der Weide and Priya Ponnapalli and Caroline Kaplan and Jiri Simsa and Shuangfeng Li and Olivier Dousse and Fan Yang and Jeff Piper and Nathan Ie and Minnie Lui and Rama Pasumarthi and Nathan Lintz and Anitha Vijayakumar and Lam Nguyen Thiet and Daniel Andor and Pedro Valenzuela and Cosmin Paduraru and Daiyi Peng and Katherine Lee and Shuyuan Zhang and Somer Greene and Duc Dung Nguyen and Paula Kurylowicz and Sarmishta Velury and Sebastian Krause and Cassidy Hardin and Lucas Dixon and Lili Janzer and Kiam Choo and Ziqiang Feng and Biao Zhang and Achintya Singhal and Tejasi Latkar and Mingyang Zhang and Quoc Le and Elena Allica Abellan and Dayou Du and Dan McKinnon and Natasha Antropova and Tolga Bolukbasi and Orgad Keller and David Reid and Daniel Finchelstein and Maria Abi Raad and Remi Crocker and Peter Hawkins and Robert Dadashi and Colin Gaffney and Sid Lall and Ken Franko and Egor Filonov and Anna Bulanova and Rémi Leblond and Vikas Yadav and Shirley Chung and Harry Askham and Luis C. Cobo and Kelvin Xu and Felix Fischer and Jun Xu and Christina Sorokin and Chris Alberti and Chu-Cheng Lin and Colin Evans and Hao Zhou and Alek Dimitriev and Hannah Forbes and Dylan Banarse and Zora Tung and Jeremiah Liu and Mark Omernick and Colton Bishop and Chintu Kumar and Rachel Sterneck and Ryan Foley and Rohan Jain and Swaroop Mishra and Jiawei Xia and Taylor Bos and Geoffrey Cideron and Ehsan Amid and Francesco Piccinno and Xingyu Wang and Praseem Banzal and Petru Gurita and Hila Noga and Premal Shah and Daniel J. Mankowitz and Alex Polozov and Nate Kushman and Victoria Krakovna and Sasha Brown and MohammadHossein Bateni and Dennis Duan and Vlad Firoiu and Meghana Thotakuri and Tom Natan and Anhad Mohananey and Matthieu Geist and Sidharth Mudgal and Sertan Girgin and Hui Li and Jiayu Ye and Ofir Roval and Reiko Tojo and Michael Kwong and James Lee-Thorp and Christopher Yew and Quan Yuan and Sumit Bagri and Danila Sinopalnikov and Sabela Ramos and John Mellor and Abhishek Sharma and Aliaksei Severyn and Jonathan Lai and Kathy Wu and Heng-Tze Cheng and David Miller and Nicolas Sonnerat and Denis Vnukov and Rory Greig and Jennifer Beattie and Emily Caveness and Libin Bai and Julian Eisenschlos and Alex Korchemniy and Tomy Tsai and Mimi Jasarevic and Weize Kong and Phuong Dao and Zeyu Zheng and Frederick Liu and Fan Yang and Rui Zhu and Mark Geller and Tian Huey Teh and Jason Sanmiya and Evgeny Gladchenko and Nejc Trdin and Andrei Sozanschi and Daniel Toyama and Evan Rosen and Sasan Tavakkol and Linting Xue and Chen Elkind and Oliver Woodman and John Carpenter and George Papamakarios and Rupert Kemp and Sushant Kafle and Tanya Grunina and Rishika Sinha and Alice Talbert and Abhimanyu Goyal and Diane Wu and Denese Owusu-Afriyie and Cosmo Du and Chloe Thornton and Jordi Pont-Tuset and Pradyumna Narayana and Jing Li and Sabaer Fatehi and John Wieting and Omar Ajmeri and Benigno Uria and Tao Zhu and Yeongil Ko and Laura Knight and Amélie Héliou and Ning Niu and Shane Gu and Chenxi Pang and Dustin Tran and Yeqing Li and Nir Levine and Ariel Stolovich and Norbert Kalb and Rebeca Santamaria-Fernandez and Sonam Goenka and Wenny Yustalim and Robin Strudel and Ali Elqursh and Balaji Lakshminarayanan and Charlie Deck and Shyam Upadhyay and Hyo Lee and Mike Dusenberry and Zonglin Li and Xuezhi Wang and Kyle Levin and Raphael Hoffmann and Dan Holtmann-Rice and Olivier Bachem and Summer Yue and Sho Arora and Eric Malmi and Daniil Mirylenka and Qijun Tan and Christy Koh and Soheil Hassas Yeganeh and Siim Põder and Steven Zheng and Francesco Pongetti and Mukarram Tariq and Yanhua Sun and Lucian Ionita and Mojtaba Seyedhosseini and Pouya Tafti and Ragha Kotikalapudi and Zhiyu Liu and Anmol Gulati and Jasmine Liu and Xinyu Ye and Bart Chrzaszcz and Lily Wang and Nikhil Sethi and Tianrun Li and Ben Brown and Shreya Singh and Wei Fan and Aaron Parisi and Joe Stanton and Chenkai Kuang and Vinod Koverkathu and Christopher A. Choquette-Choo and Yunjie Li and TJ Lu and Abe Ittycheriah and Prakash Shroff and Pei Sun and Mani Varadarajan and Sanaz Bahargam and Rob Willoughby and David Gaddy and Ishita Dasgupta and Guillaume Desjardins and Marco Cornero and Brona Robenek and Bhavishya Mittal and Ben Albrecht and Ashish Shenoy and Fedor Moiseev and Henrik Jacobsson and Alireza Ghaffarkhah and Morgane Rivière and Alanna Walton and Clément Crepy and Alicia Parrish and Yuan Liu and Zongwei Zhou and Clement Farabet and Carey Radebaugh and Praveen Srinivasan and Claudia van der Salm and Andreas Fidjeland and Salvatore Scellato and Eri Latorre-Chimoto and Hanna Klimczak-Plucińska and David Bridson and Dario de Cesare and Tom Hudson and Piermaria Mendolicchio and Lexi Walker and Alex Morris and Ivo Penchev and Matthew Mauger and Alexey Guseynov and Alison Reid and Seth Odoom and Lucia Loher and Victor Cotruta and Madhavi Yenugula and Dominik Grewe and Anastasia Petrushkina and Tom Duerig and Antonio Sanchez and Steve Yadlowsky and Amy Shen and Amir Globerson and Adam Kurzrok and Lynette Webb and Sahil Dua and Dong Li and Preethi Lahoti and Surya Bhupatiraju and Dan Hurt and Haroon Qureshi and Ananth Agarwal and Tomer Shani and Matan Eyal and Anuj Khare and Shreyas Rammohan Belle and Lei Wang and Chetan Tekur and Mihir Sanjay Kale and Jinliang Wei and Ruoxin Sang and Brennan Saeta and Tyler Liechty and Yi Sun and Yao Zhao and Stephan Lee and Pandu Nayak and Doug Fritz and Manish Reddy Vuyyuru and John Aslanides and Nidhi Vyas and Martin Wicke and Xiao Ma and Taylan Bilal and Evgenii Eltyshev and Daniel Balle and Nina Martin and Hardie Cate and James Manyika and Keyvan Amiri and Yelin Kim and Xi Xiong and Kai Kang and Florian Luisier and Nilesh Tripuraneni and David Madras and Mandy Guo and Austin Waters and Oliver Wang and Joshua Ainslie and Jason Baldridge and Han Zhang and Garima Pruthi and Jakob Bauer and Feng Yang and Riham Mansour and Jason Gelman and Yang Xu and George Polovets and Ji Liu and Honglong Cai and Warren Chen and XiangHai Sheng and Emily Xue and Sherjil Ozair and Adams Yu and Christof Angermueller and Xiaowei Li and Weiren Wang and Julia Wiesinger and Emmanouil Koukoumidis and Yuan Tian and Anand Iyer and Madhu Gurumurthy and Mark Goldenson and Parashar Shah and MK Blake and Hongkun Yu and Anthony Urbanowicz and Jennimaria Palomaki and Chrisantha Fernando and Kevin Brooks and Ken Durden and Harsh Mehta and Nikola Momchev and Elahe Rahimtoroghi and Maria Georgaki and Amit Raul and Sebastian Ruder and Morgan Redshaw and Jinhyuk Lee and Komal Jalan and Dinghua Li and Ginger Perng and Blake Hechtman and Parker Schuh and Milad Nasr and Mia Chen and Kieran Milan and Vladimir Mikulik and Trevor Strohman and Juliana Franco and Tim Green and Demis Hassabis and Koray Kavukcuoglu and Jeffrey Dean and Oriol Vinyals},
      year={2023},
      eprint={2312.11805},
      archivePrefix={arXiv},
      primaryClass={cs.CL}
}

@misc{openai2024gpt4,
      title={GPT-4 Technical Report}, 
      author={OpenAI and Josh Achiam and Steven Adler and Sandhini Agarwal and Lama Ahmad and Ilge Akkaya and Florencia Leoni Aleman and Diogo Almeida and Janko Altenschmidt and Sam Altman and Shyamal Anadkat and Red Avila and Igor Babuschkin and Suchir Balaji and Valerie Balcom and Paul Baltescu and Haiming Bao and Mohammad Bavarian and Jeff Belgum and Irwan Bello and Jake Berdine and Gabriel Bernadett-Shapiro and Christopher Berner and Lenny Bogdonoff and Oleg Boiko and Madelaine Boyd and Anna-Luisa Brakman and Greg Brockman and Tim Brooks and Miles Brundage and Kevin Button and Trevor Cai and Rosie Campbell and Andrew Cann and Brittany Carey and Chelsea Carlson and Rory Carmichael and Brooke Chan and Che Chang and Fotis Chantzis and Derek Chen and Sully Chen and Ruby Chen and Jason Chen and Mark Chen and Ben Chess and Chester Cho and Casey Chu and Hyung Won Chung and Dave Cummings and Jeremiah Currier and Yunxing Dai and Cory Decareaux and Thomas Degry and Noah Deutsch and Damien Deville and Arka Dhar and David Dohan and Steve Dowling and Sheila Dunning and Adrien Ecoffet and Atty Eleti and Tyna Eloundou and David Farhi and Liam Fedus and Niko Felix and Simón Posada Fishman and Juston Forte and Isabella Fulford and Leo Gao and Elie Georges and Christian Gibson and Vik Goel and Tarun Gogineni and Gabriel Goh and Rapha Gontijo-Lopes and Jonathan Gordon and Morgan Grafstein and Scott Gray and Ryan Greene and Joshua Gross and Shixiang Shane Gu and Yufei Guo and Chris Hallacy and Jesse Han and Jeff Harris and Yuchen He and Mike Heaton and Johannes Heidecke and Chris Hesse and Alan Hickey and Wade Hickey and Peter Hoeschele and Brandon Houghton and Kenny Hsu and Shengli Hu and Xin Hu and Joost Huizinga and Shantanu Jain and Shawn Jain and Joanne Jang and Angela Jiang and Roger Jiang and Haozhun Jin and Denny Jin and Shino Jomoto and Billie Jonn and Heewoo Jun and Tomer Kaftan and Łukasz Kaiser and Ali Kamali and Ingmar Kanitscheider and Nitish Shirish Keskar and Tabarak Khan and Logan Kilpatrick and Jong Wook Kim and Christina Kim and Yongjik Kim and Jan Hendrik Kirchner and Jamie Kiros and Matt Knight and Daniel Kokotajlo and Łukasz Kondraciuk and Andrew Kondrich and Aris Konstantinidis and Kyle Kosic and Gretchen Krueger and Vishal Kuo and Michael Lampe and Ikai Lan and Teddy Lee and Jan Leike and Jade Leung and Daniel Levy and Chak Ming Li and Rachel Lim and Molly Lin and Stephanie Lin and Mateusz Litwin and Theresa Lopez and Ryan Lowe and Patricia Lue and Anna Makanju and Kim Malfacini and Sam Manning and Todor Markov and Yaniv Markovski and Bianca Martin and Katie Mayer and Andrew Mayne and Bob McGrew and Scott Mayer McKinney and Christine McLeavey and Paul McMillan and Jake McNeil and David Medina and Aalok Mehta and Jacob Menick and Luke Metz and Andrey Mishchenko and Pamela Mishkin and Vinnie Monaco and Evan Morikawa and Daniel Mossing and Tong Mu and Mira Murati and Oleg Murk and David Mély and Ashvin Nair and Reiichiro Nakano and Rajeev Nayak and Arvind Neelakantan and Richard Ngo and Hyeonwoo Noh and Long Ouyang and Cullen O'Keefe and Jakub Pachocki and Alex Paino and Joe Palermo and Ashley Pantuliano and Giambattista Parascandolo and Joel Parish and Emy Parparita and Alex Passos and Mikhail Pavlov and Andrew Peng and Adam Perelman and Filipe de Avila Belbute Peres and Michael Petrov and Henrique Ponde de Oliveira Pinto and Michael and Pokorny and Michelle Pokrass and Vitchyr H. Pong and Tolly Powell and Alethea Power and Boris Power and Elizabeth Proehl and Raul Puri and Alec Radford and Jack Rae and Aditya Ramesh and Cameron Raymond and Francis Real and Kendra Rimbach and Carl Ross and Bob Rotsted and Henri Roussez and Nick Ryder and Mario Saltarelli and Ted Sanders and Shibani Santurkar and Girish Sastry and Heather Schmidt and David Schnurr and John Schulman and Daniel Selsam and Kyla Sheppard and Toki Sherbakov and Jessica Shieh and Sarah Shoker and Pranav Shyam and Szymon Sidor and Eric Sigler and Maddie Simens and Jordan Sitkin and Katarina Slama and Ian Sohl and Benjamin Sokolowsky and Yang Song and Natalie Staudacher and Felipe Petroski Such and Natalie Summers and Ilya Sutskever and Jie Tang and Nikolas Tezak and Madeleine B. Thompson and Phil Tillet and Amin Tootoonchian and Elizabeth Tseng and Preston Tuggle and Nick Turley and Jerry Tworek and Juan Felipe Cerón Uribe and Andrea Vallone and Arun Vijayvergiya and Chelsea Voss and Carroll Wainwright and Justin Jay Wang and Alvin Wang and Ben Wang and Jonathan Ward and Jason Wei and CJ Weinmann and Akila Welihinda and Peter Welinder and Jiayi Weng and Lilian Weng and Matt Wiethoff and Dave Willner and Clemens Winter and Samuel Wolrich and Hannah Wong and Lauren Workman and Sherwin Wu and Jeff Wu and Michael Wu and Kai Xiao and Tao Xu and Sarah Yoo and Kevin Yu and Qiming Yuan and Wojciech Zaremba and Rowan Zellers and Chong Zhang and Marvin Zhang and Shengjia Zhao and Tianhao Zheng and Juntang Zhuang and William Zhuk and Barret Zoph},
      year={2024},
      eprint={2303.08774},
      archivePrefix={arXiv},
      primaryClass={cs.CL}
}

@article{eloundou2023gpts,
  title={Gpts are gpts: An early look at the labor market impact potential of large language models},
  author={Eloundou, Tyna and Manning, Sam and Mishkin, Pamela and Rock, Daniel},
  journal={arXiv preprint arXiv:2303.10130},
  year={2023}
}

@article{zhang2023deep,
  title={Deep long-tailed learning: A survey},
  author={Zhang, Yifan and Kang, Bingyi and Hooi, Bryan and Yan, Shuicheng and Feng, Jiashi},
  journal={IEEE Transactions on Pattern Analysis and Machine Intelligence},
  year={2023},
  publisher={IEEE}
}

@inproceedings{chi2023diffusionpolicy,
	title={Diffusion Policy: Visuomotor Policy Learning via Action Diffusion},
	author={Chi, Cheng and Feng, Siyuan and Du, Yilun and Xu, Zhenjia and Cousineau, Eric and Burchfiel, Benjamin and Song, Shuran},
	booktitle={Proceedings of Robotics: Science and Systems (RSS)},
	year={2023}
}

@misc{chen2023playfusion,
      title={PlayFusion: Skill Acquisition via Diffusion from Language-Annotated Play}, 
      author={Lili Chen and Shikhar Bahl and Deepak Pathak},
      year={2023},
      eprint={2312.04549},
      archivePrefix={arXiv},
      primaryClass={cs.RO}
}

@misc{scheikl2023movement,
      title={Movement Primitive Diffusion: Learning Gentle Robotic Manipulation of Deformable Objects}, 
      author={Paul Maria Scheikl and Nicolas Schreiber and Christoph Haas and Niklas Freymuth and Gerhard Neumann and Rudolf Lioutikov and Franziska Mathis-Ullrich},
      year={2023},
      eprint={2312.10008},
      archivePrefix={arXiv},
      primaryClass={cs.RO}
}

@inproceedings{fu2024mobile,
  author    = {Fu, Zipeng and Zhao, Tony Z. and Finn, Chelsea},
  title     = {Mobile ALOHA: Learning Bimanual Mobile Manipulation with Low-Cost Whole-Body Teleoperation},
  booktitle = {arXiv},
  year      = {2024},
}

@article{brohan2023rt,
  title={Rt-2: Vision-language-action models transfer web knowledge to robotic control},
  author={Brohan, Anthony and Brown, Noah and Carbajal, Justice and Chebotar, Yevgen and Chen, Xi and Choromanski, Krzysztof and Ding, Tianli and Driess, Danny and Dubey, Avinava and Finn, Chelsea and others},
  journal={arXiv preprint arXiv:2307.15818},
  year={2023}
}

@article{brohan2022rt,
  title={Rt-1: Robotics transformer for real-world control at scale},
  author={Brohan, Anthony and Brown, Noah and Carbajal, Justice and Chebotar, Yevgen and Dabis, Joseph and Finn, Chelsea and Gopalakrishnan, Keerthana and Hausman, Karol and Herzog, Alex and Hsu, Jasmine and others},
  journal={arXiv preprint arXiv:2212.06817},
  year={2022}
}

@article{yang2023octopus,
  title={Octopus: Embodied vision-language programmer from environmental feedback},
  author={Yang, Jingkang and Dong, Yuhao and Liu, Shuai and Li, Bo and Wang, Ziyue and Jiang, Chencheng and Tan, Haoran and Kang, Jiamu and Zhang, Yuanhan and Zhou, Kaiyang and others},
  journal={arXiv preprint arXiv:2310.08588},
  year={2023}
}

@article{zhu2024language,
  title={Language-Conditioned Robotic Manipulation with Fast and Slow Thinking},
  author={Zhu, Minjie and Zhu, Yichen and Li, Jinming and Wen, Junjie and Xu, Zhiyuan and Che, Zhengping and Shen, Chaomin and Peng, Yaxin and Liu, Dong and Feng, Feifei and others},
  journal={arXiv preprint arXiv:2401.04181},
  year={2024}
}

@article{liang2023skilldiffuser,
  title={Skilldiffuser: Interpretable hierarchical planning via skill abstractions in diffusion-based task execution},
  author={Liang, Zhixuan and Mu, Yao and Ma, Hengbo and Tomizuka, Masayoshi and Ding, Mingyu and Luo, Ping},
  journal={arXiv preprint arXiv:2312.11598},
  year={2023}
}

@misc{zhen20243dvla,
      title={3D-VLA: A 3D Vision-Language-Action Generative World Model}, 
      author={Haoyu Zhen and Xiaowen Qiu and Peihao Chen and Jincheng Yang and Xin Yan and Yilun Du and Yining Hong and Chuang Gan},
      year={2024},
      eprint={2403.09631},
      archivePrefix={arXiv},
      primaryClass={cs.CV}
}

@misc{hong20233dllm,
      title={3D-LLM: Injecting the 3D World into Large Language Models}, 
      author={Yining Hong and Haoyu Zhen and Peihao Chen and Shuhong Zheng and Yilun Du and Zhenfang Chen and Chuang Gan},
      year={2023},
      eprint={2307.12981},
      archivePrefix={arXiv},
      primaryClass={cs.CV}
}

@misc{liu2024moka,
      title={MOKA: Open-Vocabulary Robotic Manipulation through Mark-Based Visual Prompting}, 
      author={Fangchen Liu and Kuan Fang and Pieter Abbeel and Sergey Levine},
      year={2024},
      eprint={2403.03174},
      archivePrefix={arXiv},
      primaryClass={cs.RO}
}

\end{document}